\documentclass{article}

\usepackage{microtype}
\usepackage{graphicx}
\usepackage{subcaption}
\usepackage{booktabs} % for professional tables

\usepackage{hyperref}
\usepackage{booktabs}
\usepackage{listings}
\usepackage{xcolor}
\usepackage{longtable}
\usepackage{array}
\usepackage[T1]{fontenc}

\lstdefinestyle{promptstyle}{
  basicstyle=\ttfamily\scriptsize,
  breaklines=true,
  breakatwhitespace=true,
  columns=fullflexible,
  frame=single,
  framesep=4pt,
  xleftmargin=2pt,
  xrightmargin=2pt,
  showstringspaces=false,
  upquote=true,
}

\usepackage[accepted]{icml2026}

\usepackage{amsmath}
\usepackage{amssymb}
\usepackage{mathtools}
\usepackage{amsthm}

\usepackage[capitalize,noabbrev]{cleveref}

\theoremstyle{plain}

\theoremstyle{definition}

\theoremstyle{remark}

\usepackage[textsize=tiny]{todonotes}

\icmltitlerunning{CareTransition-Audit: A Benchmark to Audit Discharge Summaries for Efficient Care Transitions}

\begin{document}

\twocolumn[
  \icmltitle{CareTransition-Audit: A Benchmark to Audit Discharge Summaries for Efficient Care Transitions}




  \begin{icmlauthorlist}
    \icmlauthor{Akshat Dasula}{cen,uni}
    \icmlauthor{Prasanna Desikan}{cen}
    \icmlauthor{Jaideep Srivastava}{uni}
    \icmlauthor{Shivali Dalmia}{cen}
    \icmlauthor{Abhishek Mukherji}{cen}
  \end{icmlauthorlist}

  \icmlaffiliation{uni}{Department of Computer Science \& Engineering, University of Minnesota-Twin Cities, Minneapolis, USA}
  \icmlaffiliation{cen}{Centific AI Research, Redmond, USA}

  \icmlcorrespondingauthor{Akshat Dasula}{dasul001@umn.edu}

  \icmlkeywords{Discharge summary, Clinical Documentation Audit, Large Language Models, MIMIC-IV}

  \vskip 0.3in
]



\printAffiliationsAndNotice{}  % no special notice (required even if empty)
\vspace{-15pt}
\begin{abstract}
Incomplete or inconsistent discharge documentation drives
care fragmentation and avoidable readmissions. Despite its
critical role in patient safety, auditing discharge
summaries relies on manual review and does not scale. We
propose an automated framework for auditing discharge
summaries using large language models (LLMs). Our approach
operationalizes the DISCHARGED framework into a checklist of 46 questions. Using 50 summaries from the MIMIC-IV database, with clinician ground-truth labels, we benchmark 11 LLMs. Model-assessed mean documentation completeness ranges from 54.9\% to 74.2\%, and the best-performing models achieve a Cohen's $\kappa$ values around 0.5 against clinician labels, indicating moderate agreement. All models struggle to identify ambiguous documentation (\texttt{Unclear}), highlighting a key gap in current automated auditing. This work provides a clinician-validated benchmark and zero-shot baselines for systematic quality improvement in clinical documentation.
\end{abstract}

\vspace{-20pt}

\section{Introduction}
% High-quality discharge documentation is essential for patient
% safety and a key determinant of readmission risk, facilitating
% care transitions through effective transfer of information on
% medications, follow-up plans, and prevention of adverse
% events~\cite{care, ideal}. Summaries containing key elements such
% as medication changes, pending tests, and follow-up plans are
% associated with lower 30-day readmission odds~\cite{heart}. In
% practice, physicians write the discharge note while nurses use it
% to educate patients and families; when the note is incomplete, the
% risk of miscommunication rises. Manual auditing is time-intensive
% and does not scale.

% Our work addresses this gap by formulating 46 atomic audit
% questions from the DISCHARGED framework~\cite{discharge},
% validated by a clinical expert, and applying them to 500 MIMIC-IV
% discharge summaries~\cite{mimic} restricted to surviving
% discharges. We benchmark ten LLMs in a zero-shot setting and train
% a locally deployable SFT auditor. This work contributes:
% (1)~a clinically validated 46-question evaluation framework;
% (2)~a benchmark of 500 annotated summaries for public release;
% (3)~zero-shot baselines across ten LLMs; and (4)~a
% privacy-preserving SFT auditor model.

High-quality discharge documentation is essential for patient safety and is a key determinant of readmission risk after hospitalization, as it facilitates a seamless transition from hospital to home through the effective transfer of necessary information~\cite{care, ideal} on medications and follow‑up, and preventable adverse events in the early post‑discharge period. Studies of patients hospitalized with heart failure and other high‑risk conditions have shown that discharge summaries containing key content elements such as medication changes, pending tests, and clear follow‑up plans, are associated with lower odds of 30‑day readmission \cite{heart}, showing the importance of patient‑centered documentation in safe transitions of care. In most clinical workflows, physicians write the formal discharge note while clinical nurses are responsible for educating the patient and family before discharge, using that same documentation as a reference. When the discharge note is vague or incomplete, it increases the risk of miscommunication and missed education points. Manual auditing of these notes is time and labor intensive making it difficult to perform at scale.

Our work addresses this gap by formulating 46 atomic audit questions from the DISCHARGED framework~\cite{discharge}, validated by a clinical expert, and applying them to 50 MIMIC-IV~\cite{mimic} discharge summaries restricted to surviving discharges. We benchmark eleven LLMs in a zero-shot setting and compare against clinician ground-truth labels. We make three contributions: (1)~a clinically validated 46-question evaluation framework for discharge documentation completeness; (2)~a preliminary benchmark dataset of 50 summaries with clinician-verified labels; and (3)~zero-shot baselines across eleven LLMs establishing reference performance for automated auditing.

\vspace{-5pt}
\section{Related Works}
\textbf{NLP for Clinical Documentation.}
Automated processing of clinical text has a long history in biomedical informatics. Early work focused on extracting diagnoses, medications, and adverse events from free-text EHR data using rule-based and supervised machine learning
approaches~\cite{meystre2008, mlhc}, which typically require extensive feature engineering and task-specific annotations. The introduction of transformer-based models substantially advanced the field, ClinicalBERT~\cite{alsentzer2019} adapted
bidirectional encoders to clinical corpora, while domain-specific generative models~\cite{singhal2023, nazi2024, med42v2} have demonstrated strong performance on clinical information extraction and summarization tasks. However, the predominant focus of these models has been on clinical prediction or information retrieval rather than on evaluating the documentation quality or completeness.

% Early clinical NLP focused on extracting diagnoses, medications,
% and adverse events using rule-based and supervised
% approaches~\cite{meystre2008, mlhc}. Transformer-based models
% such as ClinicalBERT~\cite{alsentzer2019} and domain-specific
% generative models~\cite{singhal2023, nazi2024, med42v2} advanced
% information extraction and summarization, but focus on clinical
% prediction rather than evaluating documentation quality.

\textbf{Discharge Summary Generation and Evaluation.} 
Most prior work focuses on automated generation of discharge summaries~\cite{llm_disch_gen, neur}, typically through fine-tuning to produce summaries that conform to predefined templates~\cite{summary} or to regenerate specific sections such as the Brief Hospital Course by removing these sections from the discharge summary and using the rest as input~\cite{ehealth, self_eval_disch_summary}. The \textit{Discharge Me!} shared task~\cite{disch_me} is an established benchmark for this, attracting contributions from multiple teams~\cite{epfl, ehealth}. However, evaluation in this line of work has relied predominantly on surface-level metrics such as ROUGE and BERTScore, or on LLM-as-a-Judge protocols~\cite{llm_judge} and expensive clinician review. Efforts to improve factual reliability, such as PDSQI-9~\cite{pdsqi} and hallucination detection methods~\cite{hallu}, address an important dimension of generation quality but remain insufficient for ensuring patient-centric completeness, as a summary can be fully factual yet still omit critical discharge elements.

% Prior work focuses on generating discharge summaries through fine-tuning to
% match templates~\cite{disch_summary_clin_guide} or regenerate
% sections like the Brief Hospital Course from the remaining
% record~\cite{ehealth, self_eval_disch_summary, llm_disch_gen, neur}.
% The \textit{Discharge Me!} benchmark~\cite{disch_me} evaluates
% section-level generation using ROUGE, BERTScore, or
% LLM-as-a-Judge~\cite{llm_judge}. Factuality methods like
% PDSQI-9~\cite{pdsqi} and hallucination detection~\cite{hallu}
% address reliability but not patient-centric completeness---a
% summary can be factual yet omit critical elements.

\textbf{Discharge Quality Frameworks and Auditing.}
Structured frameworks for discharge documentation quality such as the
AHRQ's IDEAL Framework~\cite{ideal} and DISCHARGED mnemonic framework have been proposed to enumerate elements for safe discharge documentation. Despite the availability of such frameworks, their application has remained manual and small-scale, limited by the time and labor required for clinician-led chart review~\cite{summary}. To our knowledge, no prior work has been done to audit patient-centric discharge documentation completeness for safe transitions.

% Frameworks such as AHRQ's IDEAL~\cite{ideal} and the DISCHARGED
% mnemonic~\cite{discharge} , but their application has remained manual and
% small-scale~\cite{summary}. To our knowledge, no prior work has
% operationalized a validated discharge framework into a scalable
% audit instrument with public ground-truth labels for completeness
% evaluation.

% \section{Dataset and Benchmarking Design}
% \subsection{Data Source and Cohort}
% \subsection{Operationalizing DISCHARGED Framework}
% \paragraph{Prompting Strategy.}
% Questions are divided into six prompts ($<$10 each), yielding six
% LLM calls per summary to avoid context degradation. Each uses
% indirect Chain-of-Thought~\cite{cot}: the model answers each
% question, extracts supporting evidence, and provides a
% justification. Prompts handle MIMIC-IV de-identification artifacts
% to prevent false negatives from masked fields.
% \subsection{Benchmarking}

\vspace{-10pt}
\section{Dataset and Study Design}
This work presents a retrospective audit of hospital discharge summaries using a clinically validated set of audit questions. The objective is to assess the completeness and internal consistency of discharge documentation at scale, rather than to evaluate the appropriateness of clinical care delivered. Therefore, no new clinical content is generated and no clinical outcomes are modeled.
\vspace{-6pt}
\paragraph{Data Source and Cohort.}
We use MIMIC-IV~\cite{mimic}, a publicly available, de-identified critical care database containing structured EHR data and clinical notes from Beth Israel Deaconess Medical Center. All adult inpatient admissions with an associated discharge summary were eligible for inclusion, admissions resulting in in-hospital mortality were excluded to focus on care transitions where discharge documentation directly informs downstream providers and patient education. From the eligible population, we sampled 50 discharge summaries from 50 unique patients using a stratified sampling strategy developed in consultation with
clinical experts. Stratification was performed along two axes: (1)~\textit{discharge disposition}, to ensure representation across discharge locations, and (2)~\textit{ICU utilization}, a binary indicator of whether the admission included an intensive care unit stay, capturing complexity differences between ICU and non-ICU documentation. Patient ages ranged from 23 to 91 years
($\mu = 59.5$), with a gender distribution of 56\% male and 44\% female. The mean ICU length of stay was 2.04 days and the mean admission length of stay was 6.1 days. All summaries were annotated by a clinical expert against the full set of 46 audit questions, producing clinician-verified ground-truth labels.

\subsection{Operationalizing DISCHARGED as an Audit Checklist}
We operationalize the ten components of the DISCHARGED framework~\cite{discharge} into a structured audit checklist, as shown in Table~\ref{tab:discharge_framework} and Appendix \ref{question_set}. Each question is answered using one of four labels: \texttt{Yes} if the summary explicitly contains the requested information; \texttt{No} if no relevant information is present; \texttt{Unclear} if the information is partially present but insufficiently specific due to ambiguities in clinical writing or model uncertainty; and \texttt{N/A} if the question's precondition is not met (available only for specific conditional questions). Missing documentation is interpreted strictly as a documentation gap and does not imply that the corresponding clinical care was not delivered, a distinction critical for interpreting audit results without conflating documentation quality with care quality.
\vspace{-10pt}
\begin{table}[htbp]
\centering
\caption{Component-wise Audit Questions' Structure}
\label{tab:discharge_framework}
\small
\begin{tabular}{|l|c|l|}
\hline
\textbf{Component} & \textbf{\#Q} & \textbf{Audit Focus} \\
\hline
(D)emographics          & 3 & Identity, document placement \\
(I)mportant Alerts      & 3 & Allergies, risks, precautions \\
(S)ocial Setup          & 2 & Lifestyle, social context \\
(C)omp. History         & 4 & Prior diagnoses, medications \\
(H)istory \& Exams      & 8 & Admission, vitals, exam \\
(A)ssessment \& Course  & 8 & Diagnoses, course, management \\
(R)ecorded Med $\Delta$ & 6 & Rationale, restart plans \\
(G)oals of Care         & 1 & Advance directives \\
(E)xpected Follow-up    & 3 & PCP, instructions, pending \\
(D)ischarge Info        & 8 & Date, disposition, author \\
\hline
\textbf{Total}          & \textbf{46} & \\
\hline
\end{tabular}
\end{table}
\vspace{-15pt}
\paragraph{Prompting Strategy.}
Questions are divided into six prompts ($<$10 each, see Appendix~\ref{app:prompts}) yielding six
LLM calls per summary to avoid context degradation. Each prompt employs an indirect Chain-of-Thought (CoT) strategy~\cite{cot} (asks for justification but doesn't prompt to think step-by-step), the model is instructed to answer each question with one of the designated labels, extract supporting evidence, and a brief justification linking the evidence to the label. Prompts explicitly instruct the model to recognize de-identification (e.g., masked patient identifiers) and to distinguish these from genuinely absent documentation, reducing false negatives attributable to de-identification

\vspace{-10pt}
\section{Results} \label{sec:results}

We evaluate and compare eleven LLMs against clinician labels, on 50 MIMIC-IV discharge summaries using identical prompts. Models were selected to span a range of model families and parameter scales, Gemini-3-Flash-Preview \cite{gemini}, DeepSeek v3.2 \cite{deepseek}, Phi-4 \cite{phi}, Claude Sonnet-4.5 \cite{claude}, GPT-5.4 \cite{gpt54}, GPT-4o \cite{gpt}, Grok-4.1-Fast \cite{grok}, Llama 3.3-Nemotron-49B-v1.5 \cite{nemotron}, Llama 4 Maverick \cite{llama}, and Nova-2-Lite-v1 \cite{nova} were accessed via the OpenRouter API, Qwen~2.5-7B-Instruct~\cite{qwen25} was deployed locally using HuggingFace Transformers to demonstrate the feasibility of privacy-preserving on-premise auditing. No model-specific prompt tuning or few-shot examples are used.
\vspace{-6pt}
\paragraph{Overall Agreement with Clinician Validated Labels.}

Table~\ref{tab:overall} reports agreement between each model and
the clinician across all 46 questions, with 95\% bootstrap
confidence intervals (1{,}000 resamples) for accuracy, $\kappa$,
weighted F1, and Spearman $\rho$. Claude Sonnet~4.5, Gemini~3
Flash, DeepSeek~V3, and GPT-5.4 form a top cluster with
overlapping $\kappa$ confidence intervals (range [.380, .533]),
making them statistically indistinguishable on overall agreement
despite point-estimate differences. The locally
deployed Qwen~2.5-7B ($\kappa = 0.226$) substantially
underperforms, and Phi-4's $\kappa$ CI of [.003, .089] approaches
zero, indicating its outputs are effectively uncorrelated with
clinician labels. All models remain below the $\kappa = 0.6$
threshold typically considered ``good'' agreement, indicating
that zero-shot auditing is feasible but far from solved.

% Table~\ref{tab:overall} reports agreement between each model and
% the clinician across all 46 questions. Claude Sonnet~4.5 achieves the highest Cohen's $\kappa$ (0.496)
% and GPT-5.4 and Gemini~3 Flash follow closely, while the locally
% deployed Qwen~2.5-7B ($\kappa = 0.226$) and Phi-4
% ($\kappa = 0.046$) substantially underperform. All models remain
% below the $\kappa = 0.6$ threshold typically considered ``good''
% agreement, indicating that zero-shot auditing is feasible but far
% from solved.

% ---------------------------------------------------------------
% TABLE 1: Overall model performance — MAIN RESULTS TABLE
% TODO: Update numbers when 200 GT labels are available
% ---------------------------------------------------------------
\begin{table*}[t]
\centering
\caption{Overall agreement between each LLM and the clinician on $n = 50$ MIMIC-IV discharge summaries. Values are point estimates with 95\% bootstrap confidence intervals (1{,}000 resamples). Models ranked by $\kappa$. $n$ = number of evaluated (summary, question) pairs out of $50 \times 46 = 2{,}300$.}
\label{tab:overall}
\small
\setlength{\tabcolsep}{5pt}
\begin{tabular}{lccccc}
\toprule
\textbf{Model} & \textbf{Accuracy} & \textbf{$\kappa$} & \textbf{W-F1} & \textbf{$\rho$} & \textbf{$n$} \\
\midrule
Sonnet 4.5             & .804 [.788, .821] & \textbf{.496} [.455, .533] & .815 [.800, .831] & .432 [\phantom{$-$}.178, .641] & 2168 \\
Gemini 3 Flash         & .814 [.797, .830] & .483 [.441, .520]          & .822 [.806, .838] & .537 [\phantom{$-$}.321, .706] & 2176 \\
DeepSeek V3            & .743 [.724, .761] & .423 [.388, .457]          & .775 [.759, .791] & .413 [\phantom{$-$}.146, .624] & 2176 \\
GPT-5.4                & .772 [.753, .790] & .420 [.380, .459]          & .790 [.774, .808] & .365 [\phantom{$-$}.082, .585] & 2114 \\
Nova 2 Lite            & .747 [.728, .765] & .401 [.365, .439]          & .779 [.762, .795] & .359 [\phantom{$-$}.102, .578] & 2049 \\
Nemotron 49B           & .719 [.700, .739] & .373 [.337, .410]          & .749 [.732, .765] & .474 [\phantom{$-$}.241, .656] & 2016 \\
Grok 4.1               & .721 [.702, .740] & .371 [.333, .406]          & .743 [.727, .761] & .263 [$-$.028, .517]           & 2170 \\
GPT-4o                 & .728 [.708, .745] & .370 [.333, .405]          & .760 [.742, .776] & .341 [\phantom{$-$}.064, .576] & 2176 \\
Llama 4 Maverick       & .706 [.687, .727] & .340 [.304, .378]          & .739 [.722, .757] & .335 [\phantom{$-$}.042, .581] & 2168 \\
Qwen 2.5-7B$^\dagger$  & .623 [.603, .644] & .226 [.194, .258]          & .679 [.661, .698] & .254 [$-$.011, .482]           & 2176 \\
Phi-4                  & .640 [.619, .662] & .046 [.003, .089]          & .655 [.633, .677] & .100 [$-$.196, .388]           & 2035 \\
\bottomrule
\multicolumn{6}{l}{\footnotesize $^\dagger$Locally deployed. Variation in $n$ reflects N/A labels and inference parse failures.} \\
\end{tabular}
\end{table*}
% \paragraph{Per-Label Analysis.}
% A consistent pattern emerges across all models: \texttt{Yes} labels are predicted with high precision and recall (0.80--0.94 and 0.66--0.88 respectively), while \texttt{No} achieves moderate performance (0.33--0.60 precision, 0.52--0.78 recall). The most notable finding concerns the \texttt{Unclear} label, where all models achieve near-zero precision and recall ($\leq$ 0.08 and $\leq$ 0.26). This disagreement is bidirectional, models frequently assign definitive \texttt{Yes} or \texttt{No} labels to questions that the clinician marked \texttt{Unclear}, suggesting overconfidence in resolving genuine clinical ambiguity. Conversely, models sometimes produce
% \texttt{Unclear} for questions the clinician answered definitively,
% indicating unnecessary hedging. Because the \texttt{Unclear} category captures precisely the cases where documentation is partial or ambiguous, the scenarios most likely to cause misinterpretation during care transitions, this inability of
% current models represents a key challenge for automated auditing. This finding highlights that the task requires not only information extraction but also the capacity to recognize when documented information is insufficiently specific for clinical use, a distinction that may require clinician-labeled supervision to learn reliably.
\vspace{-10pt}
\paragraph{Per-Label Analysis.}
A consistent pattern emerges across all models: \texttt{Yes} labels are predicted with high precision and recall (0.80--0.94 and 0.66--0.88 respectively), while \texttt{No} achieves moderate performance (0.33--0.60 precision, 0.52--0.78 recall). The most notable finding concerns the \texttt{Unclear} label, where all models achieve near-zero precision and recall ($\leq$ 0.08 and $\leq$ 0.26). This disagreement is bidirectional: models frequently assign definitive \texttt{Yes} or \texttt{No} labels to questions that the clinician marked \texttt{Unclear}, suggesting overconfidence in resolving genuine clinical ambiguity. Conversely, models sometimes produce \texttt{Unclear} for the questions which the clinician answered definitively, indicating unnecessary hedging. 

The clinician used \texttt{Unclear} sparingly across the full set
of labels (38 of 2{,}300, 1.7\%), with the label concentrated in
questions where partial documentation is clinically common,
pre-hospitalization functional status alone accounts for 11 of
the 38 \texttt{Unclear} labels, followed by medication restart
plans (4), and social and surgical history (3 each). A finer-grained look at these disagreements (Appendix~\ref{app:unclear}) is informative on the 38 pairs where the clinician selected \texttt{Unclear} , the two top-tier models force the choice in opposite directions,
Gemini 3 Flash labels 20 as \texttt{Yes} and 17 as \texttt{No}, while Sonnet 4.5 labels 20 as \texttt{No} and 15 as \texttt{Yes}. That two strong models systematically disagree on exactly the cases the clinician flagged as ambiguous indicates documentation ambiguity in these instances, strengthening the interpretation of \texttt{Unclear} as a real documentation signal.
Because the \texttt{Unclear} category captures precisely the cases where documentation is partial or ambiguous, the scenarios most likely to cause misinterpretation during care transitions, this inability of current models represents a key challenge for automated auditing. To understand the nature of these disagreements, our work captures free-text justifications from both the clinician and each model for every label assignment.

Ongoing analysis of these paired justifications will enable fine-grained characterization of \textit{why} models and clinicians diverge on ambiguity, whether the disagreement stems from differing interpretations of clinical language, incomplete evidence extraction, or genuine boundary cases in documentation quality, informing targeted improvements to prompting strategies and providing supervision signal for future fine-tuned auditor models.
\vspace{-12pt}
% \paragraph{Documentation Completeness.}
% Across 50 summaries, model-assessed mean completeness scores range from 54.9\% (Qwen) to 74.2\% (Gemini), confirming that substantial documentation gaps exist in MIMIC-IV discharge summaries regardless of which model performs the audit (Figure~\ref{fig:completeness}). Gemini achieves the strongest Spearman correlation with clinician completeness rankings ($\rho = 0.537$, 95\% CI [.321, .706]), while Phi-4 and Qwen show no statistically significant correlation ($p > 0.1$), meaning their summary-level scores are unreliable proxies for clinical judgment.

\paragraph{Documentation Completeness.} The clinician-rated mean
completeness across the 50 summaries is \textbf{79.0\%}, this is the primary evidence of documentation gaps in MIMIC-IV summaries.
Model-assessed completeness scores are reported as exploratory
validation of whether LLMs can recover this signal. Model-assessed
means range from 54.9\% (Qwen~2.5-7B) to 74.2\% (Gemini~3 Flash),
with all eleven models systematically \emph{under}-estimating
completeness relative to the clinician, the closest model
(Gemini) is 4.8 percentage points below, and the locally deployed
Qwen underestimates by more than 20 percentage points. This
under-estimation suggests that current zero-shot LLM auditors
are more likely to flag false positive documentation gaps than to
miss real ones, an error mode that may increase clinician alert
fatigue but is unlikely to create false confidence in incomplete
documentation. These model-assessed completeness scores should be checked alongside component-level agreement, as a model can produce a plausible aggregate completeness score while disagreeing with the clinician on which specific elements are present. Among models, Gemini~3 Flash achieves the
strongest Spearman correlation with clinician completeness
rankings ($\rho = 0.537$, 95\% CI $[.321, .706]$); Grok~4.1,
Qwen~2.5-7B, and Phi-4 show $\rho$ confidence intervals that
include zero, indicating their per-summary completeness scores
are not reliable proxies for clinician judgment.

% ---------------------------------------------------------------
% FIGURE 1: Completeness score distributions (box/violin plot)
% Show all 11 models side by side, ordered by median
% Include a horizontal dashed line for clinician GT median if avail
% ---------------------------------------------------------------
% \begin{figure}[htbp]
%     \centering
%     \includegraphics[width=1\linewidth]{heatmap.png}
%     \caption{Per-summary completeness scores (proportion of \texttt{Yes} out of 46 questions)}
%     \label{fig:completeness}
% \end{figure}
\begin{figure}[htbp]
    \centering
    \includegraphics[width=1\linewidth]{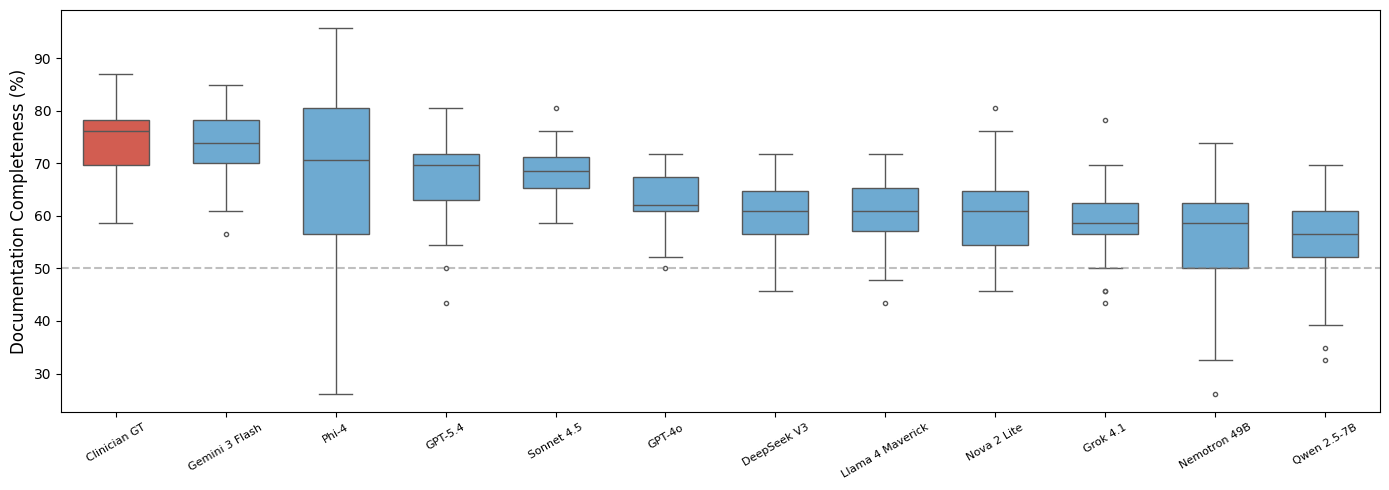}
    \caption{Per-summary completeness scores (proportion of \texttt{Yes})}
    \label{fig:completeness}
\end{figure}
\vspace{-10pt}
% \begin{figure}[htbp]
%     \centering
%     \includegraphics[width=1\linewidth]{heatmap_correct.png}
%     \caption{Cohen's $\kappa$ by DISCHARGED component and model}
%     \label{fig:heatmap}
% \end{figure}

\begin{figure*}[htbp]
    \centering
    \includegraphics[width=0.75\linewidth]{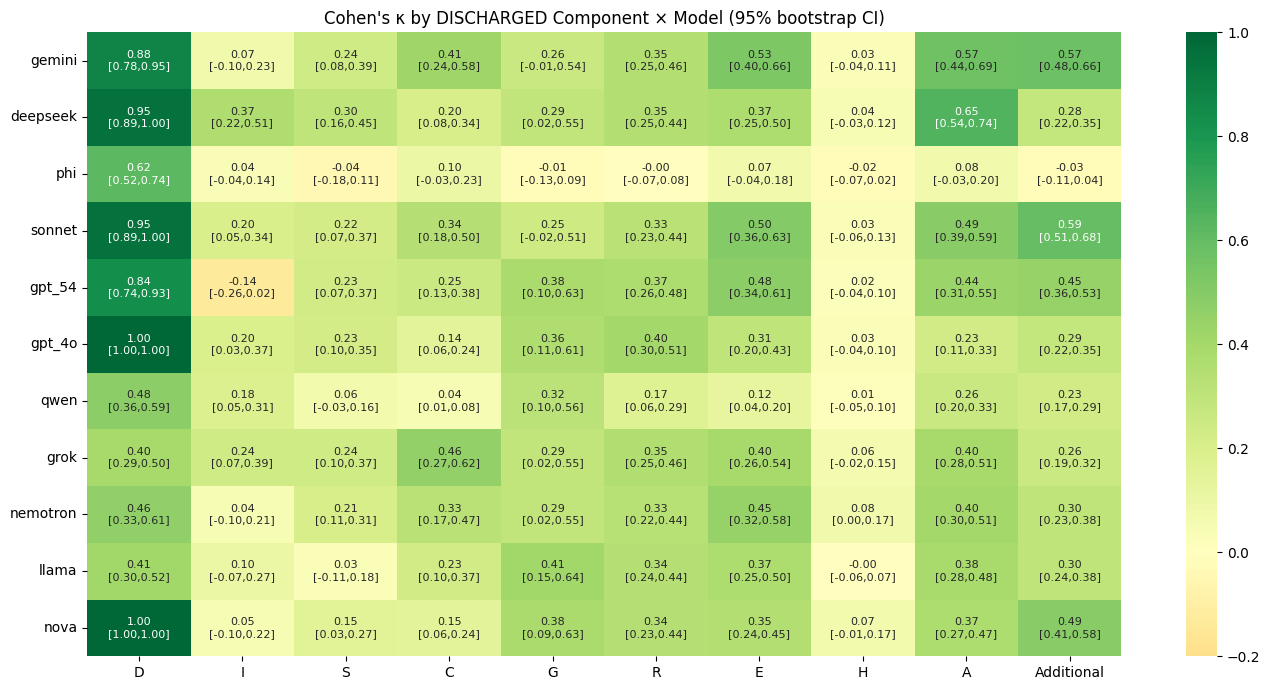}
    \caption{Cohen's $\kappa$ by DISCHARGED component and model}
    \label{fig:heatmap}
\end{figure*}

% \begin{figure}[htbp]
% \centering
% % \includegraphics[width=\columnwidth]{completeness_boxplot.pdf}
% \caption{Per-summary completeness scores (proportion of
% \texttt{Yes} out of 46 questions) across 200 summaries.
% % TODO: Generate this figure from your notebook
% }
% \label{fig:completeness}
% \end{figure}
% \vspace{-3pt}
\paragraph{Component-Level Agreement.}
Figure~\ref{fig:heatmap} shows Cohen's $\kappa$ broken down by
DISCHARGED component and model. Four distinct patterns emerge.
First, Demographics (D) shows uniformly high $\kappa$ values
(.84--1.00) with very tight CIs for most models, reflecting
ceiling-effect agreement where models and the clinician label
nearly every question \texttt{Yes}. Second, History \& Exams (H) shows the inverse
pattern, $\kappa$ near zero with CIs that span zero across all
eleven models (e.g., Sonnet $[-.06, .13]$, Gemini $[-.04, .11]$,
GPT-5.4 $[-.04, .10]$), indicating that models systematically
fail to detect documentation gaps in admission narratives.
Third, Important Alerts (I) is a less consistent failure: most
models hover near zero with wide CIs (e.g., Sonnet $[.05, .34]$,
GPT-5.4 $[-.26, .02]$), reflecting genuine difficulty with
this component. Fourth, the highest
robust agreement, where CIs are entirely positive and bounded
away from zero, appears in Assessment (A) for the top-tier
models (DeepSeek $[.54, .74]$, Sonnet $[.39, .59]$, Gemini
$[.44, .69]$) and Additional / Discharge Info (Sonnet $[.51,
.68]$, Gemini $[.48, .66]$).

\vspace{-6pt}
\section{Discussion}
% Our results show that while demographics and admission information
% are consistently documented, a substantial proportion of
% DISCHARGED elements are missing or ambiguous. The high
% \texttt{Unclear} rates across multiple audit questions indicate
% that documentation gaps are not solely a matter of omission but
% also of ambiguity---information that is partially present but
% insufficiently specific to support safe care transitions. This
% distinction is clinically significant: a medication listed without
% a documented rationale for its change, or a follow-up plan
% mentioned without a specified timeframe, can lead to
% misinterpretation by downstream providers and nursing staff who
% rely on the discharge summary for patient education and continuity
% of care.

Our framework evolved iteratively with clinical input. We initially considered auditing against the AHRQ's IDEAL framework ~\cite{ideal}. However, we realized that the discharge summary documentation does not capture the entirety of this discharge planning process. Because many IDEAL components describe clinical \textit{processes}, such as \textit{Ask open-ended questions to elicit questions and concerns of the patient}, that are not documented in discharge summaries even when performed in practice, auditing them would require evaluating clinical workflows at various points. This mismatch between process oriented guidelines and a documentation oriented evaluation would have led to unreliable audit results. The DISCHARGED framework~\cite{discharge}, designed specifically to guide summary \textit{writing}, mapped more naturally to documentation content. The initial 34 questions were a direct mapping of framework components, during pilot annotation, compound questions that evaluated multiple elements introduced labeling ambiguity. We decomposed these into 46 atomic sub-questions, reducing annotator ambiguity and improving audit granularity. Format-specific questions were excluded to maintain flexibility across the diverse documentation styles observed in clinical practice.

Clinical experts highlighted that discharge summaries are authored incrementally by multiple contributors across care transitions (e.g., ICU to floor, resident rotations), with much content added near discharge. This multi-author process can produce excessive detail in some sections and omission of critical information in others, and the level of detail varies by service, obstetric discharges are substantially more standardized than complex ICU admissions. Experts emphasized that summaries should prioritize ongoing care needs over exhaustive inpatient chronicles. Notably, nurses actively use physician-written summaries for patient education, positioning automated auditing as a practical intervention at multiple workflow stages rather than merely retrospective measurement. However, even complete documentation does not guarantee effective transitions, as downstream providers may not read the summary due to time constraints or cross-institutional barriers.
\textbf{Completeness is necessary but not sufficient for documentation quality.} The 46-item checklist measures whether the informational elements required for safe care transitions are present in a discharge summary, not whether they are well-organized, internally consistent, or clinically correct. A summary can satisfy every checklist item and still be poorly structured, contain contradictions between sections, bury critical information in narrative paragraphs, or misrepresent the clinical course. Conversely, a summary missing one or two checklist items may be highly readable and clinically sound. Our framework is therefore a screen for documentation gaps, completeness auditing is most useful as a first-pass filter that flags missing elements for clinician review, leaving questions of accuracy and clinical coherence to subsequent evaluation, whether by clinicians or by complementary frameworks such as PDSQI-9 \citep{pdsqi}, or by hallucination detection methods \citep{hallu}.
\vspace{-10pt}
\paragraph{Limitations.}
Our preliminary benchmark reflects documentation practices at a single
institution (MIMIC-IV), generalizability to other healthcare systems remains to be validated. Clinician-labeled answers reflect a single expert's assessment and would benefit from multi-annotator agreement studies to quantify inter-rater reliability. The cohort includes all surviving discharges regardless of disposition, and documentation requirements may differ across disposition types such as home versus skilled nursing facility. Results are based on 50
summaries, which may not capture the full distribution of documentation patterns across clinical services. The use of de-identified data introduces artifacts such as masked names and dates that may affect model transferability to real-world settings. \texttt{Unclear} labels represent a composite of genuine clinical
ambiguity in the documentation and the model's own uncertainty, making it difficult to disentangle documentation quality from model limitations without additional clinician justification.

\vspace{-8pt}
\section{Future Work}
Several directions emerge from this work. First, we are working with clinical experts to expand the question-set and the cohort. We also plan to recruit additional annotators for multi-annotator agreement studies. This will allow direct measurement of whether the \texttt{Unclear} category is reproducible across annotators. Second, because discharge documentation varies substantially by clinical specialty and service, further research is needed to determine the stratification strategy for different clinical contexts, enabling finer-grained analysis of documentation quality. Third, we also plan to incorporate structured MIMIC-IV data (medications, laboratory results) as supplementary auditing context,  enabling assessment of factual consistency by cross-referencing narratives against structured records, identifying cases where documented information contradicts or omits elements present in the underlying EHR data. Fourth, using the expanded clinician-labeled dataset, we aim to train a locally deployable supervised fine-tuned (SFT) auditor for real-time documentation evaluation while preserving patient privacy. This would enable scalable auditing, and serving as a real-time auditor during physician note completion or as an evaluator when an LLM generates a discharge summary.

Finally, we envision extending from documentation evaluation to documentation generation through a three-phase pipeline: (1)~longitudinal temporal EHR representation, encoding the patient's full clinical trajectory using long-context reasoning; (2)~generative LLM summarization, producing discharge summaries from the ground up rather than reformatting existing text; and (3)~auditor-based optimization, using the SFT auditor as a reward signal within a Reinforcement Learning from AI Feedback (RLAIF)~\cite{rlaif} framework. By rewarding summaries that satisfy the 46-point clinical checklist, this approach would produce documentation that is not only fluent and factually grounded but also demonstrably complete with respect to the informational elements required for safe care transitions.

% \section*{Impact Statement}
% This work aims to improve patient safety by enabling scalable evaluation of discharge documentation quality, supporting nurses and clinicians in delivering efficient patient-care and transitions. However, we emphasize that this system is a decision-support tool, not a replacement for clinical judgment. Automated audits should be reviewed by clinicians, as false negatives could create false confidence in incomplete documentation, and false positives could increase alert fatigue. 
% This work aims to improve patient safety by enabling scalable evaluation of discharge documentation quality, supporting nurses and clinicians in delivering efficient patient-care, education and transitions. However, we emphasize that this system is designed as a decision-support tool, not a replacement for clinical judgment. Automated audits should be reviewed by clinicians before acting on their outputs, as false negatives (missed gaps) could create false confidence in incomplete documentation, and false positives could increase alert fatigue.

\section*{Impact Statement}

This work aims to improve patient safety by enabling scalable
evaluation of discharge documentation quality, supporting nurses
and clinicians in delivering efficient patient care, education, and
care transitions. We emphasize that the system is designed as a
decision-support tool, not a replacement for clinical judgment.
Automated audits should be reviewed by clinicians before acting on
their outputs, as false negatives (missed gaps) could create false
confidence in incomplete documentation, and false positives could
increase alert fatigue.

All experiments use MIMIC-IV under the PhysioNet Credentialed
Health Data License, completing the required
CITI training prior to data access. API-based model evaluations
were performed through OpenRouter using its zero-data-retention
configuration, restricted to providers contractually committed to
no logging, no human review, and no use of inputs or outputs for
model training. Qwen 2.5-7B was deployed locally, demonstrating the feasibility of a fully on-premise
auditing pipeline for institutions where third-party API routing of
clinical text is not permitted. No discharge summary text or
patient-level identifiers appear in this paper or its supplementary
materials.

\bibliography{example_paper}
\bibliographystyle{icml2026}

% %%%%%%%%%%%%%%%%%%%%%%%%%%%%%%%%%%%%%%%%%%%%%%%%%%%%%%%%%%%%%%%%%%%%%%%%%%%%%%%
% %%%%%%%%%%%%%%%%%%%%%%%%%%%%%%%%%%%%%%%%%%%%%%%%%%%%%%%%%%%%%%%%%%%%%%%%%%%%%%%
% APPENDIX
%%%%%%%%%%%%%%%%%%%%%%%%%%%%%%%%%%%%%%%%%%%%%%%%%%%%%%%%%%%%%%%%%%%%%%%%%%%%%%%
%%%%%%%%%%%%%%%%%%%%%%%%%%%%%%%%%%%%%%%%%%%%%%%%%%%%%%%%%%%%%%%%%%%%%%%%%%%%%%%
\newpage
\appendix
\onecolumn

\section{Question Set}\label{question_set}
The 46 audit questions are partitioned into six prompts rather than issued as a single 46-question prompt or as 46 separate calls, balancing two competing constraints. Issuing all 46 questions in a single call risks context degradation, while issuing each question independently incurs prohibitive latency. We therefore group questions from DISCHARGED components into the same prompt, keeping each prompt under ten questions to remain within reliable context limits, and issue six LLM calls per discharge summary. The resulting partition is fixed across all models and summaries; see Table~\ref{tab:audit-checklist}.
\noindent

\begingroup
\normalsize
\setlength{\tabcolsep}{10pt}
\renewcommand{\arraystretch}{1.2}

\setlength{\LTpre}{0pt}
\setlength{\LTpost}{0pt}

\begin{longtable}{@{}r l p{8.2cm} c p{3.2cm}@{}}
\caption{Full 46-question DISCHARGED audit checklist, grouped by prompt. \texttt{N/A} is admissible only for conditional questions.}
\label{tab:audit-checklist} \\
\toprule
\textbf{\#} & \textbf{ID} & \textbf{Question} & \textbf{N/A} & \textbf{Condition} \\
\midrule
\endfirsthead

\multicolumn{5}{l}{\textit{Table~\ref{tab:audit-checklist} continued from previous page}} \\
\toprule
\textbf{\#} & \textbf{ID} & \textbf{Question} & \textbf{N/A} & \textbf{Condition} \\
\midrule
\endhead

\midrule
\multicolumn{5}{r}{\textit{continued on next page}} \\
\endfoot

\bottomrule
\endlastfoot

\multicolumn{5}{@{}l}{\textit{Prompt 1: Demographics + Important Alerts}} \\
\midrule
1  & D1   & Are basic patient demographics (age or date of birth, and sex) documented in the discharge summary? & No  & --- \\
2  & D2   & Is a patient identifier (e.g.\ name, medical record number, or patient identification number) documented, even if de-identified? & No  & --- \\
3  & D3   & Is patient contact information (e.g.\ address or phone number) documented, even if de-identified or blank? & No  & --- \\
4  & I1   & Is the patient's allergy status documented (either specific allergies listed, or an explicit statement such as NKDA/NDA/no known allergies)? & No  & --- \\
5  & I2   & If specific allergies are listed, are the allergens and their reaction types (e.g.\ rash, anaphylaxis) documented? & Yes & Patient has no allergies \\
6  & I3   & Are any other clinical alerts documented, such as adverse drug reactions, special risks, or precautions? & No  & --- \\
\midrule
\multicolumn{5}{@{}l}{\textit{Prompt 2: Social Setup + Comprehensive History + Goals of Care}} \\
\midrule
7  & S1   & Does the discharge summary document any social history (e.g.\ smoking status, alcohol use, substance use, occupation, or living situation)? & No  & --- \\
8  & S2   & Does the discharge summary describe the patient's pre-hospitalization functional status (e.g.\ independence, mobility, baseline exercise tolerance)? & No  & --- \\
9  & C1   & Does the discharge summary state the patient's past medical history (e.g.\ previous diagnoses or chronic conditions)? & No  & --- \\
10 & C2   & Does the discharge summary state the patient's past surgical history? & Yes & Explicit ``no prior surgeries'' \\
11 & C3   & Is a pre-admission medication list documented? & No  & --- \\
12 & C4   & If a pre-admission medication list is documented, does it include doses and frequencies (not just drug names)? & Yes & C3 = No \\
13 & G1   & Is there any documentation of goals of care, advance directives, code status, or advance care planning? & No  & --- \\
\midrule
\multicolumn{5}{@{}l}{\textit{Prompt 3: Recorded Medication Changes + Expected Follow-up}} \\
\midrule
14 & R1   & Is a discharge medication list documented? & No  & --- \\
15 & R2   & If a discharge medication list is documented, does it include the purpose or indication for each medication? & Yes & R1 = No \\
16 & R3   & If a discharge medication list is documented, does it include dose, route, and/or frequency information? & Yes & R1 = No \\
17 & R4   & Are any medication changes (new medications started, medications stopped, or dose adjustments) clearly documented? & No  & --- \\
18 & R5   & For documented medication changes, is the specific clinical rationale for each change provided? & Yes & R4 = No \\
19 & R6   & For medications stopped during the stay, is there a clear plan for whether or when they should be restarted? & Yes & No medications stopped \\
20 & E1   & Are follow-up instructions or appointments included in the discharge summary? & No  & --- \\
21 & E2   & Are there clear instructions regarding which outstanding investigations or pending results need to be reviewed or traced in the outpatient setting? & No  & --- \\
22 & E3   & Is the contact information for the Primary Care Provider (PCP) listed in the summary, even if de-identified or blank? & No  & --- \\
\midrule
\multicolumn{5}{@{}l}{\textit{Prompt 4: History \& Examinations}} \\
\midrule
23 & H1   & Does the discharge summary document the reason for the patient's admission? & No  & --- \\
24 & H2   & Does the discharge summary mention the admission date? & No  & --- \\
25 & H3   & Does the discharge summary document the source of referral or mode of admission (e.g.\ self-referral, ED, transfer from another facility)? & No  & --- \\
26 & H4   & Does the discharge summary document vital signs or clinical parameters on presentation? & No  & --- \\
27 & H5   & Does the discharge summary document targeted physical examination findings on presentation? & No  & --- \\
28 & H6   & Is the presenting symptom characterized with any detail (e.g.\ nature, onset, duration, progression, alleviating/exacerbating factors)? & No  & --- \\
29 & H7   & Are associated symptoms or significant negatives (especially to rule out red-flag symptoms) documented? & No  & --- \\
30 & H8   & Is relevant surgical history, drug history, or family history documented where pertinent to the presenting complaint? & No  & --- \\
\midrule
\multicolumn{5}{@{}l}{\textit{Prompt 5: Assessment \& Clinical Course}} \\
\midrule
31 & A1   & Are medical diagnoses given in the summary (actual medical diagnosis, not just symptoms)? & No  & --- \\
32 & A2   & Is the severity or complication level of the main diagnoses clearly described (e.g.\ KDIGO stage for AKI)? & No  & --- \\
33 & A3   & Where appropriate, does the summary include a brief one-sentence problem representation explaining the key features that support the diagnosis? & No  & --- \\
34 & A4   & Are clinical investigations listed (i.e.\ blood tests, lab tests, imaging, diagnostic procedures)? & No  & --- \\
35 & A5   & Is there a concise description of the patient's hospital course or clinical trajectory during admission? & No  & --- \\
36 & A6   & Does the summary describe the management plan for each main problem, including conservative measures, pharmacologic treatments, and any procedures or surgeries? & No  & --- \\
37 & A7   & Is the response to treatment documented for each major problem (e.g.\ resolution of symptoms, improvement in oxygen requirement, trending of creatinine)? & No  & --- \\
38 & A8   & If recommended investigations or treatments were withheld or stopped, is the reason documented (e.g.\ patient preference, goals of care, futility, risk\,$>$\,benefit)? & Yes & None withheld or stopped \\
\midrule
\multicolumn{5}{@{}l}{\textit{Prompt 6: Discharge Information (Additional)}} \\
\midrule
39 & Add1 & Is the date of discharge documented? & No  & --- \\
40 & Add2 & Is the specialty of the doctor that discharged the patient included in the summary? & No  & --- \\
41 & Add3 & Is the discharge disposition documented (e.g.\ home, rehab, skilled nursing facility, step-down care)? & No  & --- \\
42 & Add4 & Is the type of discharge documented (e.g.\ normal, against medical advice, abscondment)? & No  & --- \\
43 & Add5 & Is the condition of the patient at discharge described (e.g.\ stable, improved, critical)? & No  & --- \\
44 & Add6 & Is hospital contact information listed for patient perusal, even if de-identified or blank? & No  & --- \\
45 & Add7 & Is information about the discharge summary writer included, even if de-identified? & No  & --- \\
46 & Add8 & Is the attending physician or discharging provider identified in the summary, even if de-identified? & No  & --- \\

\end{longtable}
\endgroup
% \twocolumn

\section{Prompts}
\label{app:prompts}

All six prompts share a common structure: an auditor persona statement scoped to the prompt's components, a note about de-identified content, a rule set governing the four allowed labels (\texttt{Yes}, \texttt{No}, \texttt{Unclear}, \texttt{N/A}), the audit questions for that prompt, and a strict JSON output schema requiring an answer, verbatim evidence from the discharge summary, and a brief justification for each question. The discharge summary text is appended at the end of each prompt. All prompts are issued zero-shot, with no system message and no model-specific tuning. The placeholder \texttt{\{discharge\_summary\}} is replaced at runtime with the verbatim discharge summary text. The full audit checklist with conditional logic is given in Table~\ref{tab:audit-checklist}.

\subsection{Prompt 1: Demographics + Important Alerts}

\begin{lstlisting}[style=promptstyle]
You are a clinical documentation auditor who works on demographic information and patient alerts.

You will be given a discharge summary. Your task is to answer the following audit questions based ONLY on the information present in the discharge summary.

Note:
- You are working with a de-identified dataset, information maybe explicitly stated but the details of it maybe blank (e.g. contact information)
- Give justification clearly when dealing with information which has blanks or dashes

Rules:
- Do NOT infer or assume information.
- Answers must be strictly one of: "Yes", "No", "Unclear", or "N/A".
- Use "Unclear" ONLY if partial or ambiguous information is present.
- If the information is completely absent, answer "No".
- Use "N/A" ONLY when the question's precondition does not apply (e.g., a conditional question whose triggering condition is not met).
- Evidence must be a direct quote or exact phrase(s) from the discharge summary.
- Justification must briefly explain why the evidence supports the selected answer.
- Do NOT add any content outside the specified JSON structure.


Audit Questions for Demographic Information:
1) Are basic patient demographics (age or date of birth, and sex) documented in the discharge summary?

2) Is a patient identifier (e.g. name, medical record number, or patient identification number) documented, even if de-identified?

3) Is patient contact information (e.g. address or phone number) documented, even if de-identified or blank?

Audit Questions for Important Alerts:
1) Is the patient's allergy status documented (either specific allergies listed, or an explicit statement such as NKDA/NDA/no known allergies)?

2) If specific allergies are listed, are the allergens and their reaction types (e.g. rash, anaphylaxis) documented? Answer "N/A" if the patient is documented as having no
allergies.

3) Are any other clinical alerts documented, such as adverse drug reactions, special risks, or precautions?

--------------------

Output Format (STRICT - valid JSON only):

{
  "D": {
    "1": {
      "answer": "Yes/No/Unclear",
      "evidence": "Exact quoted text or Not documented",
      "justification": "Brief explanation linking the evidence to the answer"
    },
    "2": {
      "answer": "Yes/No/Unclear",
      "evidence": "Exact quoted text or Not documented",
      "justification": "Brief explanation linking the evidence to the answer"
    },
    "3": {
      "answer": "Yes/No/Unclear",
      "evidence": "Exact quoted text or Not documented",
      "justification": "Brief explanation linking the evidence to the answer"
    }
  },
  "I": {
    "1": {
      "answer": "Yes/No/Unclear",
      "evidence": "Exact quoted text or Not documented",
      "justification": "Brief explanation linking the evidence to the answer"
    },
    "2": {
      "answer": "Yes/No/Unclear/N/A",
      "evidence": "Exact quoted text or Not documented",
      "justification": "Brief explanation linking the evidence to the answer"
    },
    "3": {
      "answer": "Yes/No/Unclear",
      "evidence": "Exact quoted text or Not documented",
      "justification": "Brief explanation linking the evidence to the answer"
    }
  }
}

-------------------------
Discharge Summary:
{discharge_summary}
\end{lstlisting}

\subsection{Prompt 2: Social Setup + Comprehensive History + Goals of Care}

\begin{lstlisting}[style=promptstyle]
You are a clinical documentation auditor who works on social history, past medical history, and goals-of-care documentation.

You will be given a discharge summary. Your task is to answer the following audit questions based ONLY on the information present in the discharge summary.

Note:
- You are working with a de-identified dataset, information maybe explicitly stated but the details of it maybe blank (e.g. contact information)
- Give justification clearly when dealing with information which has blanks or dashes

Rules:
- Do NOT infer or assume information.
- Answers must be strictly one of: "Yes", "No", "Unclear", or "N/A".
- Use "Unclear" ONLY if partial or ambiguous information is present.
- If the information is completely absent, answer "No".
- Use "N/A" ONLY when the question's precondition does not apply (e.g., a conditional question whose triggering condition is not met).
- Evidence must be a direct quote or exact phrase(s) from the discharge summary.
- Justification must briefly explain why the evidence supports the selected answer.
- Do NOT add any content outside the specified JSON structure.


Audit Questions for Social Set up:
1) Does the discharge summary document any social history (e.g. smoking status, alcohol use, substance use, occupation, or living situation)?

2) Does the discharge summary describe the patient's pre-hospitalization functional status (e.g. whether they lived independently, mobility level, baseline exercise tolerance)?

Audit Questions for Comprehensive Past Med History:
1) Does the discharge summary state the patient's past medical history (e.g. previous diagnoses or chronic conditions)?

2) Does the discharge summary state the patient's past surgical history? Answer "N/A" if there is an explicit statement that the patient has no prior surgeries.

3) Is a pre-admission medication list documented?

4) If a pre-admission medication list is documented, does it include doses and frequencies (not just drug names)? Answer "N/A" if no pre-admission medication list is present.


Audit Questions for Goals-of-care documentation:
1) Is there any documentation of goals of care, advance directives, code status, or advance care planning (e.g. serious illness conversations, advance medical directives)?

--------------------

Output Format (STRICT - valid JSON only):

{
  "S": { "1": {...}, "2": {...} },
  "C": { "1": {...},
         "2": { "answer": "Yes/No/Unclear/N/A", ... },
         "3": {...},
         "4": { "answer": "Yes/No/Unclear/N/A", ... } },
  "G": { "1": {...} }
}

-------------------------
Discharge Summary:
{discharge_summary}
\end{lstlisting}

\subsection{Prompt 3: Recorded Medication Changes + Expected Follow-up}

\begin{lstlisting}[style=promptstyle]
You are a clinical documentation auditor who works on medication changes and follow-up instructions.

You will be given a discharge summary. Your task is to answer the following audit questions based ONLY on the information present in the discharge summary.

Note:
- You are working with a de-identified dataset, information maybe explicitly stated but the details of it maybe blank (e.g. contact information)
- Give justification clearly when dealing with information which has blanks or dashes

Rules:
- Do NOT infer or assume information.
- Answers must be strictly one of: "Yes", "No", "Unclear", or "N/A".
- Use "Unclear" ONLY if partial or ambiguous information is present.
- If the information is completely absent, answer "No".
- Use "N/A" ONLY when the question's precondition does not apply (e.g., a conditional question whose triggering condition is not met).
- Evidence must be a direct quote or exact phrase(s) from the discharge summary.
- Justification must briefly explain why the evidence supports the selected answer.
- Do NOT add any content outside the specified JSON structure.


Audit Questions for Record of Medication Changes:
1) Is a discharge medication list documented?

2) If a discharge medication list is documented, does it include the purpose or indication for each medication? Answer "N/A" if no discharge medication list is present.

3) If a discharge medication list is documented, does it include dose, route, and/or frequency information? Answer "N/A" if no discharge medication list is present.

4) Are any medication changes (new medications started, medications stopped, or dose adjustments) clearly documented?

5) For medication changes that are documented, is the specific clinical rationale for each change provided? Answer "N/A" if no medication changes are documented.

6) For medications stopped during the stay, is there a clear plan for whether or when they should be restarted? Answer "N/A" if no medications were stopped.

Audit Questions for Expected Follow-up instructions:
1) Are follow up instructions or appointments included in the discharge summary?

2) Are there clear instructions regarding which outstanding investigations or pending results need to be reviewed or traced in the outpatient setting?

3) Is the contact information for the Primary Care Provider (PCP) listed in the summary, even if de-identified or blank?

--------------------

Output Format (STRICT - valid JSON only):

{
  "R": { "1": {...},
         "2": { "answer": "Yes/No/Unclear/N/A", ... },
         "3": { "answer": "Yes/No/Unclear/N/A", ... },
         "4": {...},
         "5": { "answer": "Yes/No/Unclear/N/A", ... },
         "6": { "answer": "Yes/No/Unclear/N/A", ... } },
  "E": { "1": {...}, "2": {...}, "3": {...} }
}

-------------------------
Discharge Summary:
{discharge_summary}
\end{lstlisting}

\subsection{Prompt 4: History \& Examinations}

\begin{lstlisting}[style=promptstyle]
You are a clinical documentation auditor who works on history of presenting complaint and physical examination findings.

You will be given a discharge summary. Your task is to answer the following audit questions based ONLY on the information present in the discharge summary.

Note:
- You are working with a de-identified dataset, information maybe explicitly stated but the details of it maybe blank (e.g. contact information)
- Give justification clearly when dealing with information which has blanks or dashes

Rules:
- Do NOT infer or assume information.
- Answers must be strictly one of: "Yes", "No", or "Unclear".
- Use "Unclear" ONLY if partial or ambiguous information is present.
- If the information is completely absent, answer "No".
- Evidence must be a direct quote or exact phrase(s) from the discharge summary.
- Justification must briefly explain why the evidence supports the selected answer.
- Do NOT add any content outside the specified JSON structure.


Audit Questions for History of presenting complaint and
physical examination findings:
1) Does the discharge summary document the reason for the patient's admission?

2) Does the discharge summary mention the admission date?

3) Does the discharge summary document the source of referral or mode of admission (e.g. self-referral, emergency department, transfer from another facility)?

4) Does the discharge summary document vital signs or clinical parameters on presentation?

5) Does the discharge summary document targeted physical examination findings on presentation?

6) Is the presenting symptom characterized with any detail (e.g. nature, onset, duration, progression, alleviating/exacerbating factors)?

7) Are associated symptoms or significant negatives (especially to rule out red-flag symptoms) documented?

8) Is relevant surgical history, drug history, or family history documented where pertinent to the presenting complaint (e.g. risk factors affecting pretest probability or differential diagnosis)?

--------------------

Output Format (STRICT - valid JSON only):

{ "H": { "1": {...}, "2": {...}, "3": {...}, "4": {...},
         "5": {...}, "6": {...}, "7": {...}, "8": {...} } }

-------------------------
Discharge Summary:
{discharge_summary}
\end{lstlisting}

\subsection{Prompt 5: Assessment \& Clinical Course}

\begin{lstlisting}[style=promptstyle]
You are a clinical documentation auditor who works on assessment and clinical course.

You will be given a discharge summary. Your task is to answer the following audit questions based ONLY on the information present in the discharge summary.

Note:
- You are working with a de-identified dataset, information maybe explicitly stated but the details of it maybe blank (e.g. contact information)
- Give justification clearly when dealing with information which has blanks or dashes

Rules:
- Do NOT infer or assume information.
- Answers must be strictly one of: "Yes", "No", "Unclear", or "N/A".
- Use "Unclear" ONLY if partial or ambiguous information is present.
- If the information is completely absent, answer "No".
- Use "N/A" ONLY when the question's precondition does not apply (e.g., a conditional question whose triggering condition is not met).
- Evidence must be a direct quote or exact phrase(s) from the discharge summary.
- Justification must briefly explain why the evidence supports the selected answer.
- Do NOT add any content outside the specified JSON structure.


Audit Questions for Assessment & Clinical Course:
1) Are medical diagnoses given in the summary (actual medical diagnosis, not just symptoms)?

2) Is the severity or complication level of the main diagnoses clearly described (e.g., KDIGO stage for AKI)?

3) Where appropriate, does the summary include a brief one-sentence problem representation explaining the key features that support the diagnosis?

4) Are clinical investigations listed (i.e blood tests, lab tests, imaging, diagnostic procedures)?

5) Is there a concise description of the patient's hospital course or clinical trajectory during admission?

6) Does the summary describe the management plan for each main problem, including conservative measures, pharmacologic treatments, and any procedures or surgeries?

7) Is the response to treatment documented for each major problem (e.g., resolution of symptoms, improvement in oxygen requirement, trending of creatinine)?

8) If recommended investigations or treatments were withheld or stopped, is the reason documented (e.g., patient preference, goals of care, futility, risk greater than benefit)? Answer "N/A" if no investigations or treatments appear to have been withheld or stopped.

--------------------

Output Format (STRICT - valid JSON only):

{ "A": { "1": {...}, "2": {...}, "3": {...}, "4": {...},
         "5": {...}, "6": {...}, "7": {...},
         "8": { "answer": "Yes/No/Unclear/N/A", ... } } }

-------------------------
Discharge Summary:
{discharge_summary}
\end{lstlisting}

\subsection{Prompt 6: Discharge Information (Additional)}

\begin{lstlisting}[style=promptstyle]
You are a clinical documentation auditor who works on general discharge documentation completeness.

You will be given a discharge summary. Your task is to answer the following audit questions based ONLY on the information present in the discharge summary.

Note:
- You are working with a de-identified dataset, information maybe explicitly stated but the details of it maybe blank (e.g. contact information)
- Give justification clearly when dealing with information which has blanks or dashes

Rules:
- Do NOT infer or assume information.
- Answers must be strictly one of: "Yes", "No", or "Unclear".
- Use "Unclear" ONLY if partial or ambiguous information is present.
- If the information is completely absent, answer "No".
- Evidence must be a direct quote or exact phrase(s) from the discharge summary.
- Justification must briefly explain why the evidence supports the selected answer.
- Do NOT add any content outside the specified JSON structure.


Audit Questions for Additional:
1) Is the date of discharge documented?

2) Is the specialty of the doctor that discharged the patient included in the summary?

3) Is the discharge disposition documented (e.g. discharged home, rehab, skilled nursing facility, step-down care)?

4) Is the type of discharge documented (e.g. normal, against medical advice, abscondment)?

5) Is the condition of the patient at discharge described (e.g. stable, improved, critical)?

6) Is hospital contact information listed for patient perusal, even if de-identified or blank?

7) Is information about the discharge summary writer included, even if de-identified?

8) Is the attending physician or discharging provider identified in the summary, even if de-identified?

--------------------

Output Format (STRICT - valid JSON only):

{ "Additional": {
    "1": {...}, "2": {...}, "3": {...}, "4": {...},
    "5": {...}, "6": {...}, "7": {...}, "8": {...} } }

-------------------------
Discharge Summary:
{discharge_summary}
\end{lstlisting}

\section{Distribution of Clinician \texttt{Unclear} Labels}
\label{app:unclear}

Across the 50 annotated discharge summaries and 46 audit
questions, the clinician applied the \texttt{Unclear} label 38
times (1.7\% of all labels). Table~\ref{tab:unclear-distribution}
reports the per-question distribution. The label is heavily
concentrated: three questions account for 18 of the 38
\texttt{Unclear} labels, and 14 of the 46 questions received no
\texttt{Unclear} labels in any summary.

Several patterns emerge. The single most ambiguous question is
pre-hospitalization functional status (S2), which is rarely
documented as a discrete field in MIMIC-IV summaries.
Medication-related ambiguity in the Recorded Medication
Changes (R) component (questions R2, R3, R4, R5, R6 collectively
contribute 9 \texttt{Unclear} labels), reflecting that even when
medications are listed, the surrounding details required by the
audit (purpose, dose/route/frequency, change rationale, restart
plans) are inconsistently provided. The long tail of single-hit
questions across the remaining components indicates that
\texttt{Unclear} is otherwise rare.

\begin{table}[htbp]
\centering
\caption{Per-question distribution of clinician \texttt{Unclear}
labels across 50 summaries. Questions not appearing in the table
received zero \texttt{Unclear} labels.}
\label{tab:unclear-distribution}
\small
\setlength{\tabcolsep}{6pt}
\renewcommand{\arraystretch}{1.15}
\begin{tabular}{@{}clc@{}}
\toprule
\textbf{ID} & \textbf{Question (abbreviated)} & \textbf{Count} \\
\midrule
S2   & Pre-hospitalization functional status            & 11 \\
R6   & Restart plan for stopped medications             & 4  \\
S1   & Social history documented                        & 3  \\
C2   & Past surgical history documented                 & 3  \\
R5   & Clinical rationale for medication changes        & 2  \\
Add4 & Type of discharge documented                     & 2  \\
I2   & Allergens and reaction types documented          & 2  \\
A5   & Hospital course / clinical trajectory described  & 1  \\
Add3 & Discharge disposition documented                 & 1  \\
R3   & Discharge med list: dose/route/frequency         & 1  \\
R2   & Discharge med list: purpose/indication           & 1  \\
Add5 & Patient's condition at discharge described       & 1  \\
E3   & PCP contact information listed                   & 1  \\
E2   & Outstanding investigations flagged for follow-up & 1  \\
R4   & Medication changes clearly documented            & 1  \\
H6   & Presenting symptom characterized with detail     & 1  \\
A2   & Severity/complication of main diagnoses          & 1  \\
A8   & Reason documented for withheld/stopped treatment & 1  \\
\midrule
\multicolumn{2}{r}{\textbf{Total}} & \textbf{38} \\
\bottomrule
\end{tabular}
\end{table}

Table~\ref{tab:unclear-model-labels} reports how three
representative models, the two top performers by overall
$\kappa$ (Gemini 3 Flash, Sonnet 4.5) and the locally deployed
Qwen 2.5-7B, labeled the 38 questions on which the clinician
selected \texttt{Unclear}. First, Gemini and Sonnet almost never
reproduce the clinician's \texttt{Unclear} label (1 and 2 cases
respectively), instead forcing a definitive Yes/No on
genuinely ambiguous documentation. Critically, the two models
resolve this ambiguity in \emph{opposite directions} on the same
38 pairs, Gemini selects Yes 20 times while Sonnet selects No
20 times. That two top-tier models disagree systematically on
clinician-flagged ambiguous cases is itself evidence that the
underlying documentation is genuinely ambiguous. Second, Qwen 2.5-7B hedges substantially more readily (10 \texttt{Unclear} and 5 \texttt{N/A} out of 38),
suggesting that smaller locally-deployed models may exhibit the
opposite failure mode from the API tier. Further analysis is needed on these findings  as follow-up work.

\begin{table}[h]
\centering
\caption{Model label distributions on the 38 (summary, question)
pairs where the clinician labeled \texttt{Unclear}. Gemini and
Sonnet rarely produce \texttt{Unclear}, splitting their forced
choices in opposite directions on the same pairs; Qwen hedges
far more readily.}
\label{tab:unclear-model-labels}
\small
\setlength{\tabcolsep}{8pt}
\renewcommand{\arraystretch}{1.15}
\begin{tabular}{@{}lcccc@{}}
\toprule
\textbf{Model} & \textbf{Yes} & \textbf{No} & \textbf{Unclear} & \textbf{N/A} \\
\midrule
Gemini 3 Flash       & 20 & 17 & 1  & 0 \\
Sonnet 4.5           & 15 & 20 & 2  & 0 \\
Qwen 2.5-7B$^\dagger$ & 10 & 13 & 10 & 5 \\
\bottomrule
\multicolumn{5}{l}{\footnotesize $^\dagger$Locally deployed.} \\
\end{tabular}
\end{table}

\end{document}